\documentclass[sigconf]{acmart}
\AtBeginDocument{%
  \providecommand\BibTeX{{%
    Bib\TeX}}}

\copyrightyear{2026}
\acmYear{2026}
\setcopyright{cc}
\setcctype{by}
\acmConference[SEAMS '26]{21st International Conference on Software Engineering for Adaptive and Self-Managing Systems}{April 13--14, 2026}{Rio de Janeiro, Brazil}
\acmBooktitle{21st International Conference on Software Engineering for Adaptive and Self-Managing Systems (SEAMS '26), April 13--14, 2026, Rio de Janeiro, Brazil}
\acmPrice{}
\acmDOI{10.1145/3788550.3794868}
\acmISBN{979-8-4007-2445-9/2026/04}

\usepackage{graphicx}
\usepackage{textcomp}
\usepackage{xcolor}
\def\BibTeX{{\rm B\kern-.05em{\sc i\kern-.025em b}\kern-.08em
    T\kern-.1667em\lower.7ex\hbox{E}\kern-.125emX}}
    
\usepackage{multirow}

\usepackage{xspace}
\usepackage{amsfonts}
\usepackage{colortbl}
\usepackage{amsmath}
\usepackage{algorithm}
\usepackage{subcaption}
\usepackage{enumitem}
\usepackage{mathrsfs}
\usepackage{url}

\newcommand{\code}[1]{{\small\texttt{#1}}}

\newtheorem{theorem}{Theorem}
\newtheorem{definition}[theorem]{Definition}

\begin{document}

\title{\oursystem: Causality-based Self-Adaptation for Autonomous IoT Systems}

\author{Houssam Hajj Hassan}
\email{houssam.hajj_hassan@telecom-sudparis.eu}
\affiliation{\institution{Télécom SudParis, Institut Polytechnique de Paris}
\country{France}}
\author{Ajay Kattepur}
\email{ajay.kattepur@ericsson.com}
\affiliation{\institution{Ericsson Research Artificial Intelligence}
\country{India}}
\author{Denis Conan}
\email{denis.conan@telecom-sudparis.eu}
\affiliation{\institution{Télécom SudParis, Institut Polytechnique de Paris}
\country{France}}
\author{Georgios Bouloukakis}
\email{gbouloukakis@upatras.gr}
\affiliation{\institution{Department of Electrical and Computer Engineering, University of Patras}\country{Greece}}
\affiliation{\institution{Télécom SudParis, Institut Polytechnique de Paris}\country{France}}

\renewcommand{\shortauthors}{Hajj Hassan et al.}

\newcommand{\oursystem}{\textsc{CRAFTER}\xspace}

\newcommand{\vLocations}[1][]{\mathcal{L}}
\newcommand{\vLocation}[1][i]{{l}_{#1}}
\newcommand{\vRateOfLocation}[1][]{\lambda_{\vLocation}}

\newcommand{\vDevices}[1][]{\mathcal{D}}
\newcommand{\vDevice}[1][j]{d_{#1}}
\newcommand{\vObservation}[1][j]{o_{#1}}
\newcommand{\vTopicOfDevice}[1][]{T_{\vDevice}}
\newcommand{\vRateOfDevice}[1][]{\lambda_{\vDevice}}

\newcommand{\vVirtualSensor}[1][j]{vs_{#1}}
\newcommand{\vVirtualSensors}[1][]{VS}
\newcommand{\vResources}[1][]{\rho}
\newcommand{\vSubTopicsOfVirtualSensor}[1][]{T_{\vVirtualSensor}^{sub}}
\newcommand{\vPubTopicsOfVirtualSensor}[1][]{T_{\vVirtualSensor}^{pub}}
\newcommand{\vVirtualSensorProcessing}[1][]{\tau_{\vVirtualSensor}}

\newcommand{\vApplications}[1][]{A}
\newcommand{\vApplication}[1][i]{a_{#1}}
\newcommand{\vApplicationQoS}[1][i]{\vApplication.qos}
\newcommand{\vTopicsOfApp}[1][]{T_{\vApplication}}
\newcommand{\vMetaverse}[1][i]{{a}^{meta}_{#1}}

\newcommand{\vEdgeNodes}[1][]{E}
\newcommand{\vEdgeNode}[1][i]{e_{#1}}
\newcommand{\vEdgeCpu}[1][i]{\vEdgeNode.proc}
\newcommand{\vEdgeMem}[1][i]{\vEdgeNode.mem}
\newcommand{\vEdgeDisk}[1][i]{\vEdgeNode.disk}
\newcommand{\vNodeUtilization}[1][]{U(\vEdgeNode)}

\newcommand{\vRouter}[1][i]{\mathcal{X}_{#1}}
\newcommand{\vLinkBandwidth}[1][i]{w_{#1}}
\newcommand{\vLinkDelay}[1][i]{\Delta_{#1}}
\newcommand{\vLinkLossRate}[1][i]{\omega_{#1}}

\newcommand{\vLatency}[1][]{\delta}
\newcommand{\vThroughput}[1][]{\Theta}
\newcommand{\vLossRate}[1][]{\omega}
\newcommand{\vSystemProperties}[1][]{\mu_{qos}}
\newcommand{\vContextProperties}[1][]{\mu_{context}}

\begin{abstract}
  This paper presents \oursystem, an automated framework for designing and deploying self-adaptive IoT systems using Causal Reinforcement Learning (CRL). As IoT devices increasingly populate pervasive computing spaces, smart environments are enabled with advanced monitoring and interactive services. The dynamic nature of these environments, such as fluctuating workloads and evolving application demands, poses significant challenges in maintaining consistent Quality of Service (QoS) levels of IoT applications. While existing self-adaptation techniques offer adaptive capabilities, they are often designed to deal with specific application domains, hindering the design of self-adaptive solutions that can be re-used across multiple IoT verticals.
In addition, there is a lack of automated pipelines that act on identifying key performance drivers to take effective adaptation decisions.
\oursystem addresses these issues by using \textit{Causality} as a formal framework for performance analysis of IoT systems. \oursystem generates causal graphs to uncover dependencies among system components and guide adaptation decisions based on cause–effect relationships. Then, adaptation agents can leverage this knowledge to take more effective adaptation decisions in dynamic situations. Our experimental evaluation demonstrates how CRAFTER enables deriving causal graphs spanning diverse IoT use cases. Furthermore, we showcase how CRAFTER improves self-adaptation performance by 25\% compared to state-of-the-art Reinforcement Learning-based approaches.

\end{abstract}

\ccsdesc[300]{Computer systems organization~Sensor networks}
\ccsdesc[300]{Computer systems organization~Reconfigurable computing}
\ccsdesc[500]{Computing methodologies~Reinforcement learning}
\ccsdesc[500]{Computing methodologies~Causal reasoning and diagnostics}
\ccsdesc[500]{Computing methodologies~Intelligent agents}

\keywords{Self-adaptive systems, IoT, Causal discovery, Reinforcement Learning}

\maketitle

\section{Introduction}
\label{chap:crafter/sec:intro}
In today's pervasive computing environments, Internet of Things (IoT) devices enable smarter spaces that offer enhanced monitoring and interactive services.
Such IoT services and applications have become mainstream within multiple systems, such as smart homes, smart agriculture, or smart building automation.
With the increasing integration of IoT sensors within consumer devices, a new generation of devices has emerged: virtual, augmented and mixed reality (XR). Software and inter-networking of these XR devices, along with advances in Edge computing and novel networking technologies (6G) have led to the emergence of \textit{Metaverse} applications~\cite{metaverse-acm}.

The Metaverse provides a realistic way to reflect real-world environments that include famous landmarks (Eiffel tower, Statue of Liberty) or places that are familiar to individuals (home, school) through Digital Twin (DT) replicas.
For this purpose, Metaverse applications rely on IoT devices to capture real-time data from physical spaces, which is then synchronized with virtual environments to create immersive and interactive experiences.
The integration of these applications with physical environments relies on advanced IoT platforms consisting of devices that sense physical phenomena and generate data. This data is then processed, filtered, or aggregated before being used by applications~\cite{ream}.
Given this complexity, the performance of IoT applications largely depends on the ability of IoT systems to meet stringent Quality of Service (QoS) requirements, such as low latency, high availability, and accuracy.

However, modern smart spaces are inherently dynamic and unpredictable. Changes in the number of IoT devices, network conditions, and available computational resources create a continuously evolving environment. For example, during major public events such as concerts or sports games in a large-scale smart city infrastructure, the influx of thousands of visitors can cause a significant surge in demand for IoT-enabled services such as real-time traffic management, public safety surveillance, and smart parking systems. This sudden spike in data traffic and sensor activity can result in network congestion and resource bottlenecks at Edge servers.
To ensure that IoT systems can effectively adapt to changes while maintaining strict QoS requirements, it is essential to leverage self-adaptation techniques within these systems. Self-adaptive solutions monitor system performance and respond to changes, adjusting configurations in real-time to avoid performance degradation. Several existing approaches~\cite{aquacep, ream, distream, planiot, adaptation-survey, 9239654, 9101225} have explored self-adaptive mechanisms for IoT systems, focusing on monitoring QoS metrics like network congestion and resource utilization. However, these solutions often fall short by neglecting the broader context in which IoT systems operate. They typically rely on reactive measures, failing to predict and prevent issues before they lead to QoS violations~\cite{adaptation-survey}.
In addition, the majority of these works do not offer formal approaches for investigating the causes of QoS violations; adaptation decisions are often solely based on changes in QoS metrics~\cite{aquacep, planiot, distream, adaptation-survey}.
Nevertheless, context information provides valuable insights into \textit{when} and \textit{how} to take adaptation decisions. 
More importantly, adaptation logic in prior work is typically tailored to specific application domains, limiting reuse and requiring domain expertise.

This paper introduces \oursystem, a causality-based self-adaptation solution for autonomous IoT systems.
The core contribution of \oursystem lies in its application-independent approach for causal graph generation, enabling the identification of context and performance variables that influence system behavior. This enables creating a semi-automated pipeline that allows IoT system developers to follow the same approach for designing self-adaptation solutions approaches across diverse environments, whether in smart cities, homes, or industrial settings.
In addition, the causal graph generation process requires minimal domain expertise, reducing human intervention in the self-adaptation pipeline.

To enable the design of such an automated pipeline, \oursystem first relies on Causality ~\cite{causality-primer} as a theoretical framework for the analysis of IoT systems' performance.
This enables defining a formal methodology for identifying causal relationships among the different variables in a smart space.
\oursystem then integrates Causal models with decision-making approaches such as Reinforcement Learning (RL)~\cite{rl} to take informed adaptation decisions that account for both QoS and contextual factors affecting performance.
\oursystem enhances the reliability of modern autonomous IoT systems through the following key contributions:
\begin{itemize}[leftmargin=*]
    \item [--] A semi-automated approach for designing self-adaptive IoT systems across heterogeneous smart environments.
    \item [--] A formal methodology for performance analysis using Causal Models.
    \item [--] An automated Causal RL (CRL) pipeline to take informed adaptation decisions in dynamic environments.
    \item[--] An open-source prototype implementation showing CRAFTER's effectiveness in self-adaptive IoT systems.
\end{itemize}

The remainder of the paper is organized as follows. Section~\ref{chap:crafter/sec:overview} presents an overview of \oursystem starting from a motivating use case to highlight the need for autonomous adaptation in IoT environments.
Section~\ref{chap:crafter/sec:iotmodel} formally presents the \oursystem IoT system model.
We detail our solution in Section~\ref{chap:crafter/sec:approach}, and provide an experimental evaluation of \oursystem in Section~\ref{chap:crafter/sec:experiments}.
In Section~\ref{chap:crafter/sec:related}, we provide an overview of related work.
Section~\ref{chap:crafter/sec:limitations} goes over the limitations of \oursystem, and Section~\ref{chap:crafter/sec:conclusion} concludes the paper with future directions.

\section{The \oursystem Approach}
\label{chap:crafter/sec:overview}

We first highlight the challenges in maintaining QoS requirements of applications in dynamic environments in Section~\ref{subsec:louvre-scenario} through a motivating use case.
We then illustrate in Section~\ref{subsec:challenges} the challenges designing generic self-adaptation solutions across different IoT domains.
Finally, we provide an overview of how \oursystem addresses these challenges in Section~\ref{subsec:overview}.

\subsection{Le~Louvre Scenario}
\label{subsec:louvre-scenario}

\begin{figure}[t]
  \centering
  \includegraphics[width=\linewidth]{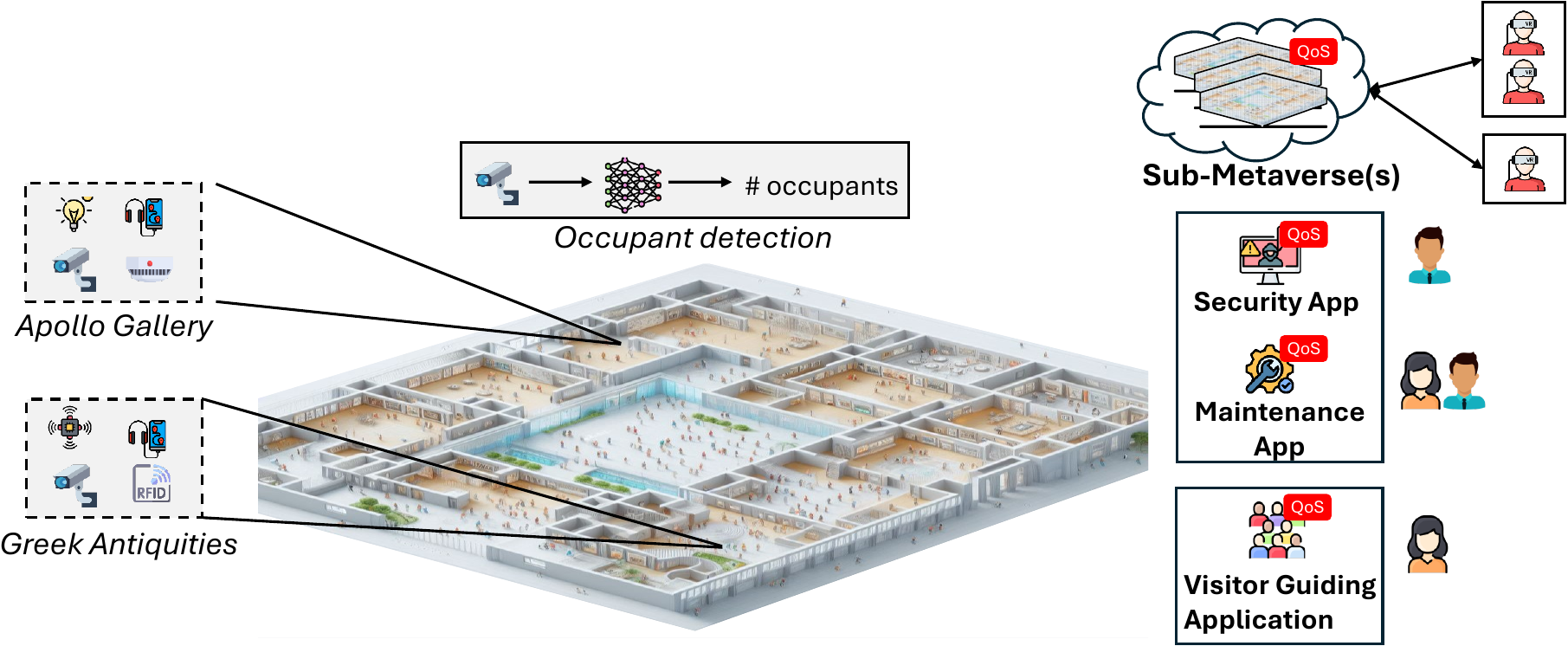}
    \vspace{-0.15in}
  \caption{Smart museum scenario}
 \vspace{-0.25in}
   \Description{An image showing sensor, virtual sensor, and application deployments in a smart museum.}
  \label{fig:use-case}
\end{figure}

To illustrate the complexity and challenges involved in designing self-adaptive IoT systems, we consider a representative example: the Louvre Museum in Paris.
This scenario exemplifies the operation of a large-scale, dynamic IoT environment where diverse devices and applications are deployed to serve multiple stakeholders, each with distinct QoS requirements.
In addition, this scenario demonstrates the challenges associated with low-latency requirements and how \oursystem addresses such challenges.
\newline
Le~Louvre is divided into three sectors (Richelieu, Sully, and Denon), and each sector comprises several exhibition rooms that contain sculptures, artifacts, and paintings spanning various civilizations and cultural movements.
Le~Louvre administrators wish that everyone, including people not physically present in Paris, is able to explore works of art displayed in the museum and participate in interactive events.
For this purpose, a DT replica of Le~Louvre is created and deployed in the Metaverse, where remote visitors (\textit{meta-occupants}) are able to visit Le~Louvre virtually.
The DT replica (\textit{Le~Louvre sub-metaverse}) captures \textit{static} properties of Le~Louvre including the organization of sectors, and placements of works of art.
Additionally, to provide a realistic and immersive experience, Le~Louvre sub-metaverse should accurately reflect the \textit{dynamic} conditions of Le~Louvre in real-time.
For this purpose, IoT devices are used to provide: (i)~data from light sensors for a realistic lighting conditions, (ii)~noise data for a realistic acoustic experience, (iii)~ visitors' movements to reflect crowd conditions in exhibition rooms, and (iv)~location of artifacts (e.g., temporary exhibitions).

As shown in Figure~\ref{fig:use-case}, to provide such monitoring and interactive capabilities, IoT devices  are deployed to monitor artifacts (motion sensors, RFID tags), visitors (cameras), and ambient conditions (temperature/humidity sensors, smoke detectors).
Device data may additionally be processed by \textit{Virtual Sensors} before being consumed by applications~\cite{ream}.
For example, video frames may be analyzed and aggregated with motion and WiFi data to provide more accurate estimations of the occupancy in each exhibition room.
Finally, applications receive data to provide services for different stakeholders.
For instance, Le~Louvre security personnel may use a dashboard that shows monitoring information in different sectors of the museum, and visitors may use a dedicated application for checking waiting times in different exhibition rooms.

\subsection{Challenges in Designing Portable and Generic Self-adaptation Solutions}
\label{subsec:challenges}

The Louvre scenario illustrates the complexity of deploying self-adaptive IoT applications in dynamic, multi-stakeholder environments. While many approaches for deploying self-adaptive systems exist in the literature, they are often designed in an ad hoc, domain-specific manner, making them difficult to generalize or reuse across different application domains (e.g., smart museums, factories, or cities)~\cite{abbas2023self}. We identify two key limitations in existing approaches for designing self-adaptive IoT systems:
\begin{enumerate}[leftmargin=*]
    \item Designing effective self-adaptation logic today still requires significant domain expertise~\cite{weyns2022deep, zavala2018sacre}. Developers must identify  -- either manually or with the help of a domain expert -- which context and performance variables are most relevant, understand their relationships, and encode specific adaptation strategies. This process is often manual, time-consuming, and difficult to replicate across different domains. 
    \item Existing self-adaptive systems lack a systematic and reusable approach for identifying what to adapt and when~\cite{gheibi2021applying, abbas2023self}. Most solutions rely on reactive, rule-based mechanisms or machine learning models trained on domain-specific datasets. These approaches often fail to generalize or provide explainability.
\end{enumerate}

Given these challenges, there is a clear need for self-adaptation frameworks that are not only capable of responding to dynamic environments, but that also facilitate re-usability and that are generalizable across domains. \oursystem addresses this gap through Causal Reinforcement Learning techniques to automate the design of adaptation logic and enable its reuse across deployments. We provide next an overview of \oursystem's high-level architecture.

\begin{figure}[t]
  \centering
  \includegraphics[width=\linewidth]{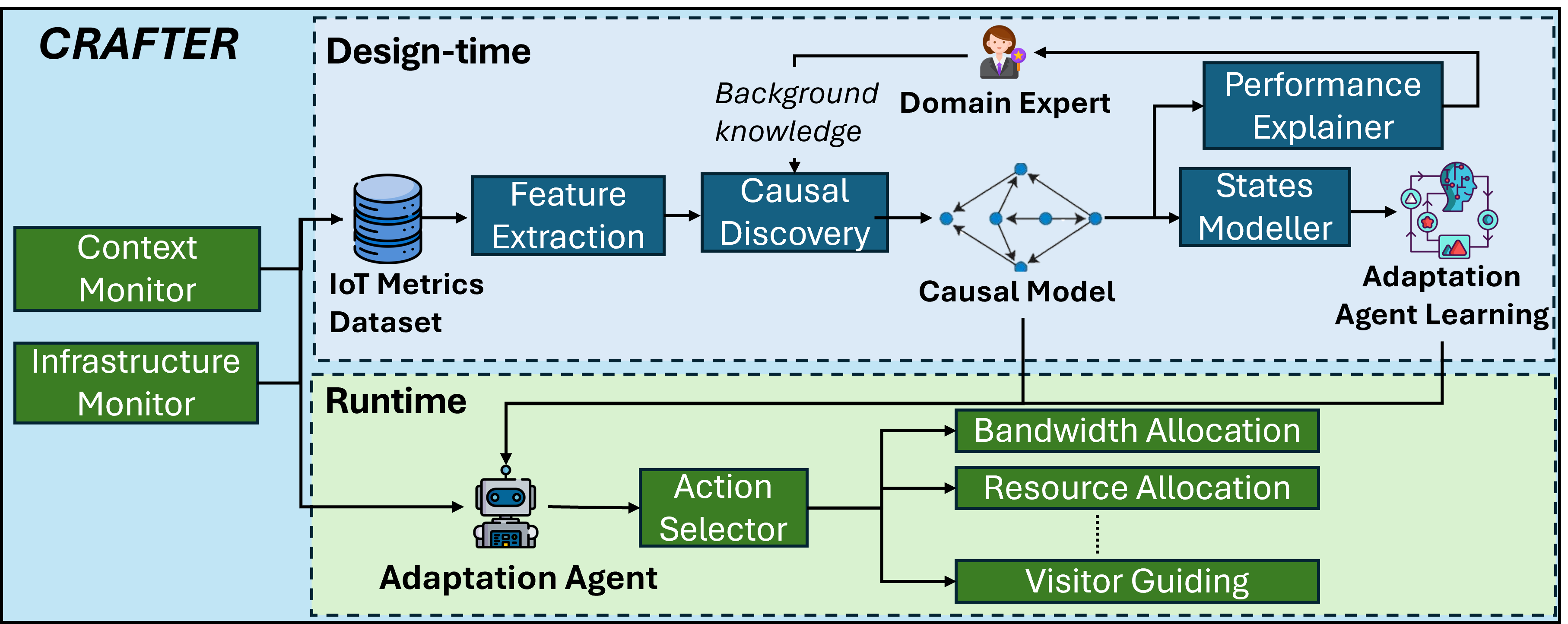}
  \caption{\oursystem's high-level architecture}
  \Description{The architecture of CRAFTER, consisting of a design-tie phase and a runtime phase.}
  \label{fig:architecture}
\end{figure}

\begin{figure}[t]
  \centering
  \includegraphics[width=\linewidth]{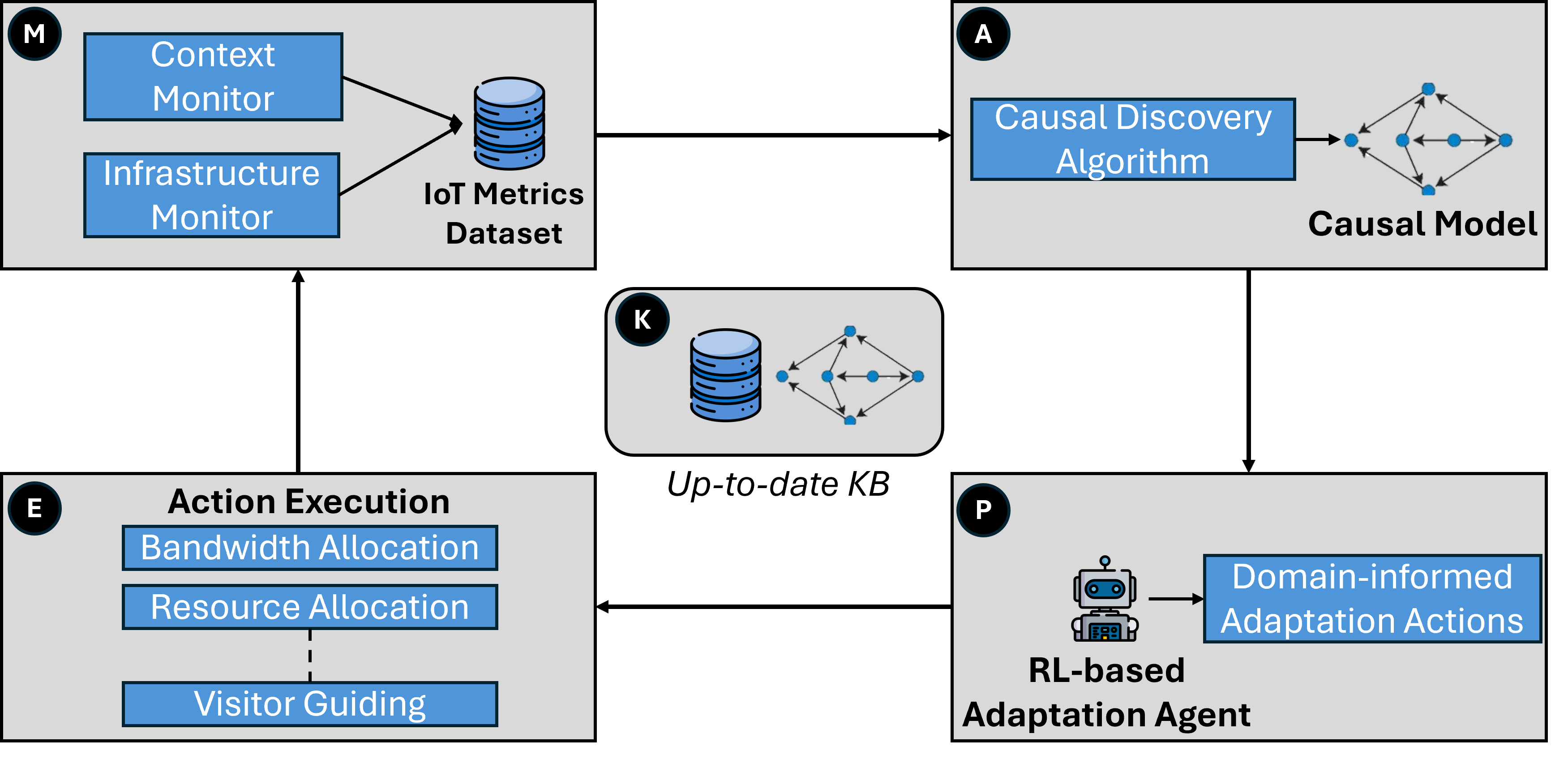}
  \caption{\oursystem's integration within the MAPE-K loop}
 \vspace{-0.2in}
 \Description{CRAFTER components inside the MAPE-K loop.}
  \label{fig:mapek-integration}
\end{figure}

\subsection{\oursystem Overview}
\label{subsec:overview}

\oursystem offers causality-based self-adaptive capabilities for modern IoT systems to enhance their autonomy, reliability, and ensure that QoS levels of applications are maintained, even in dynamic environments.
This is achieved by identifying causal relationships among variables that affect the performance of the IoT system.
This knowledge drives more efficient decision-making at runtime.

To achieve this, Figure~\ref{fig:architecture} shows how \oursystem is exploited in two phases: (1) at \textit{design-time} for its deployment in smart environments; and (2) at \textit{runtime} for enabling autonomous and efficient decision-making of adaptation agent.
Figure~\ref{fig:mapek-integration} shows \oursystem's integration within the MAPE-K adaptation loop~\cite{kephart2003vision}.

\paragraph{Design-time phase} \oursystem can be deployed in any IoT environment at design-time to extract causal relationships within the IoT system. 
For this purpose, an \textit{IoT metrics dataset} is required: this can be realistic historical data collected through prior monitoring of the smart environment, and/or synthetic datasets generated with DT approaches or ML-based dataset generation techniques~\cite{edict, wang2023sudokusens}.
For a more comprehensive understanding of the system behavior, it is important that the dataset is also enriched with context information, such as date/time of day, location of devices, number of occupants, and other contextual information that is relevant to the smart environment.
Such context information allows the generation of richer causal graphs, and subsequently enables adaptation agents to take better quality adaptation actions. 
However, note that this is not mandatory and we do not require any additional processing or labeling of the dataset.
\oursystem accepts as input regular datasets that can be collected from monitoring platforms deployed in smart spaces (in \texttt{csv} format, which requires minimal processing).
Then, \textit{Causal Models} are generated through \textit{Causal Discovery}~\cite{causal-survey}; these represent the dependencies among the different features (e.g., resource consumption, number of visitors) captured in the dataset (more details are provided in Section~\ref{subsec:causality}).
To obtain more accurate causal graphs, domain knowledge may be incorporated through expert elicitation, and/or IoT representations of smart spaces, such as Ontologies~\cite{brick, bot, saref}.
The obtained causal model serves a dual purpose.
First, the graph is used by the \textit{Performance Explainer} to explain the behavior of the IoT system to domain experts by showing causal relationships that can enhance business decisions or enable better tuning of configuration parameters for improved QoS of applications.
Second, the causal graph enables efficient self-adaptation in dynamic situations by identifying the leading causes of performance degradation, including both QoS metrics (e.g., resource consumption) and context information (number of occupants, day and time, etc.).
These are then used by the \textit{States Modeller} to model the most important states to be observed by the \textit{Adaptation Agent}.

\paragraph{Runtime phase.}
At runtime, the deployed IoT system and its applications become autonomous, guided by a generated causal graph that enables precise and effective adaptation actions. Unlike traditional RL models that may react to changes only when QoS metrics change, the causal graph allows the \textit{Adaptation Agent} to identify and act on key factors that lead to such degradation. By observing states defined by the causal relationships, the \textit{Adaptation Agent} can take preemptive and more efficient actions through the \textit{Action Selector}, addressing potential issues before they manifest as performance problems.
For instance, in Le~Louvre scenario, a higher number of occupants leads to a higher workload on processing nodes.
When this relationship is learned, the \textit{Adaptation Agent} can monitor the number of occupants as a leading indicator of potential QoS degradation. Instead of waiting for processing nodes to become overloaded, which could lead to higher latency or data loss, \oursystem enables early adaptation by reconfiguring resources as the number of occupants increases. This automated approach reduces the likelihood of QoS degradation, ensuring that applications' QoS levels are maintained in changing conditions.
The causal graph also helps in resolving hidden dependencies that might not be immediately apparent, allowing the system to adjust in ways that would otherwise be overlooked by purely reactive methods. For instance, if an increase in noise level is found to be an indicator of a high number of occupants, the system can proactively increase available resources to avoid performance degradation. 

\paragraph{\oursystem's integration within the MAPE-K loop} Figure~\ref{fig:mapek-integration} shows how \oursystem components fit within the different phases of the MAPE-K loop~\cite{kephart2003vision}. The Monitor (M) phase includes the collection of the IoT metrics dataset through IoT monitoring platforms. 
This dataset is then used in the Analyze (A) phase by causal discovery algorithms to generate the causal graphs.
The IoT metrics dataset and the generated causal graph constitute the Knowledge Base (KB -- K), and are continuously updated to account for changes.
The Planning (P) phase involves the RL agent taking adaptation actions in response to changing conditions of the environments, and the Execute (E) phase applies the adaptation actions to the Managed System (i.e., the IoT environment).

\section{Modelling Dynamic IoT Environments}
\label{chap:crafter/sec:iotmodel}
This section presents a formal description of target IoT systems for which \oursystem is designed, followed by a formal definition of the self-adaptive QoS problem in the IoT.

\subsection{IoT System Formal Model}
\label{subsec:iot-formal-model}
We rely on the publish/subscribe architecture~\cite{eugster2003many} to model data exchange between the components of the IoT system.
For this purpose, a dedicated message broker is deployed for receiving, filtering, and forwarding messages based on topic-based subscription filters. 

We consider that a smart space is divided into various locations $\vLocations$, where each location $\vLocation \in \vLocations$ contains a set of deployed IoT devices $\vDevices(\vLocation)$.
A device $\vDevice$ provides one (or more) observation $\vObservation$ capturing a physical property that is specific to its location.
For example, a smoke detector provides an observation related to ``smoke'' in ``exhibition hall 1''.
$\vDevice$ publishes data to one or more topics $\vTopicOfDevice$; we compose topics by coupling the observation $\vObservation$ with the location $\vLocation$ where the device is deployed.
Hence, a temperature sensor that is deployed in ``exhibition hall 3'' publishes data to the topic \code{hall3/temperature}.
Devices either generate data periodically (e.g., temperature sensors), or in an event-driven manner (e.g., a motion sensor) following a probability distribution (e.g., normal distribution).
The data generation rate of each device is defined as $\vRateOfDevice$. Therefore, the total traffic generated by all devices in a location is the sum of the individual rates, denoted as $\vRateOfLocation = \sum \lambda_{\vDevices(\vLocation)}$.
Note that this traffic can be collected and processed from \oursystem components to create an IoT metrics dataset $\textbf{D}$.

Devices in the same location connect to a shared network switch
$\vRouter$ that has specific QoS characteristics such as bandwidth $\vLinkBandwidth$, link delay $\vLinkDelay$ and loss rate $\vLinkLossRate$.
To enhance sensing capabilities, a set of \textit{virtual sensors} $\vVirtualSensors$ is deployed to process data from IoT devices and produce high-level observations.
For example, motion and camera data may be combined to create a new ``room occupancy'' observation.
For this purpose, a set of dedicated Edge nodes $\vEdgeNodes$ are available for hosting virtual sensors.
Each edge node $\vEdgeNode \in \vEdgeNodes$ offers resources related to processing power, memory, and disk space; we denote these characteristics by the tuple $\vResources(\vEdgeNode) =(\vEdgeCpu, \vEdgeMem, \vEdgeDisk)$.
Each virtual sensor $\vVirtualSensor \in \vVirtualSensors$ subscribes to a set of topics $\vSubTopicsOfVirtualSensor$ to receive and process relevant data, and publish new observations to $\vPubTopicsOfVirtualSensor$.
The processing time of a virtual sensor $\vVirtualSensorProcessing$ depends on the rate at which it receives data, the processing algorithm/technique used, and the available resources of the Edge node where $\vVirtualSensor$ is deployed.
To offer monitoring capabilities and interactive services to the smart space administrators and occupants, a set of applications $\vApplications$ are deployed.
Each application $\vApplication \in \vApplications$ subscribes to a set of topics $\vTopicsOfApp$ to receive data of interest from devices and virtual sensors.
Since applications may be used by different stakeholders, an application defines a set of requirements $\vApplicationQoS$ representing QoS metrics that need to be satisfied.
For instance, a security dashboard application specifies that the tail latency $\vLatency_{tail}$ should be less than $2~seconds$, and the minimum throughput $\vThroughput_{min}$ greater than $4~Mbps$.
As described in Section~\ref{subsec:louvre-scenario}, a DT replica of the smart space may also be available for meta-occupants to virtually join the space.
We denote such applications by $\vMetaverse \in \vApplications^{meta}$.

\subsection{The Self-Adaptive QoS Management Problem}
\label{subsec:formal-adaptation}
To ensure correct application behavior, space administrators typically configure their systems to sustain the required QoS levels.
However, as smart spaces are dynamic environments, multiple changes usually occur that affect the QoS requirements of applications.
These can be changes in $\vRateOfLocation$, the utilization of Edge resources $\vNodeUtilization$, or changing QoS requirements $\vApplicationQoS$ and thus, self-adaptive solutions should be deployed to adapt to these changes.
Hence, for an application $\vApplication$ with QoS requirements $\vApplicationQoS = ( \vLatency_{max}, \vThroughput_{min}, \vLossRate_{max} )$, the following conditions must be satisfied at any time $t$:
\begin{equation}
    \vLatency \leq \vLatency_{max}; \; \; \vThroughput \geq \vThroughput_{min}; \; \; \vLossRate \leq \vLossRate_{max} \qquad \forall t
\end{equation}
Typically, to detect performance issues, a set of metrics are continuously monitored to provide insights into the performance of the system~\cite{adaptation-survey}.
In addition to the QoS requirements of applications, this set contains (i)~network characteristics ($\vLinkBandwidth$, $\vLinkDelay$, $\vLinkLossRate$, \dots); (ii)~resource usage of Edge nodes $\vNodeUtilization$; (iii)~number of devices $|\vDevices|$ and applications $|\vApplications|$.
We refer to this set of monitored QoS metrics by $\vSystemProperties$~\cite{adaptation-survey}.
Usually, when changes in $\vSystemProperties$ cross a threshold, adaptation actions are executed manually or through an automated process~\cite{adaptation-survey}.
However, when such changes are detected, this often implies that QoS violations already occurred.
This happens because $\vSystemProperties$ are chosen arbitrarily or through experts' domain knowledge.

To solve these issues, \oursystem relies on a formal methodology for identifying key variables driving the performance of the IoT system and affecting the QoS requirements of applications.
In particular, \oursystem leverages Causal Discovery for identifying, alongside $\vSystemProperties$, \textit{context} variables (denoted by $\vContextProperties$) that contribute to fluctuations of the IoT system performance.
Such variables may include day, time, weather, location, number of occupants in location, noise levels, and other context information found to affect the system performance.
Monitoring $\vContextProperties$ in addition to $\vSystemProperties$ contributes to the early detection of QoS violations.
The identified variables empower the decision-making process of the Adaptation Agent that is able to take more efficient adaptation actions.
The next section provides further details about this process.

\begin{table}
\caption{Parameters notations in \oursystem's IoT model.}
\centering {\small {\renewcommand{\arraystretch}{1.2}
\scalebox{0.8}{
\begin{tabular}{|c|c|c|c|}
\hline
\textbf{Notation} & \textbf{Description} & \multicolumn{1}{l|}{\textbf{Notation}} & \multicolumn{1}{l|}{\textbf{Description}} \\ \hline
$\vLocation \in \vLocations$ & locations & $\vEdgeNode \in \vEdgeNodes$ & Edge nodes \\ \hline
$\vDevice \in \vDevices(\vLocation)$ & devices in $\vLocation$ & $\vResources(\vEdgeNode)$ & Edge Resources \\ \hline
$\vObservation$ & observations from $\vDevice$ & $\vVirtualSensor \in \vVirtualSensors$ & virtual sensors \\ \hline
$\vTopicOfDevice$ & $\vDevice$'s topics & $\vSubTopicsOfVirtualSensor; \vPubTopicsOfVirtualSensor$ & $\vVirtualSensor$'s topics \\ \hline
$\vRateOfDevice$ & $\vDevice$'s publication rate & $\vVirtualSensorProcessing$ & $\vVirtualSensor$'s processing delay \\ \hline
$\vRateOfLocation$ & data traffic per $\vLocation$ & $\vApplication \in \vApplications$ & applications \\ \hline
$\vLinkBandwidth$ & bandwidth & $\vTopicsOfApp$ & $\vApplication$'s topics \\ \hline
$\vLinkDelay$ & link delay & $( \vLatency, \vThroughput, \vLossRate)$ & latency, throughput, loss rate \\ \hline
$\vLinkLossRate$ & loss rate & $(\vSystemProperties, \vContextProperties)$ & system, context properties \\ \hline
\end{tabular}}}}
\label{tab:my-table}
\Description{Table including notation of parameters in the CRAFTER IoT model.}
\end{table}

\section{Enabling Causality-based Adaptation}
\label{chap:crafter/sec:approach}
\oursystem leverages the causality framework~\cite{causality-primer} as a formal theoretical methodology for identifying cause-and-effect relationships in IoT systems through observational data, including context information.
More specifically, based on an IoT metrics dataset $\textbf{D}$, 
causal models are generated to identify $\vSystemProperties$ and $\vContextProperties$.
These are then used as drivers for informed decision-making.

\subsection{Identifying Causal Relationships in IoT Systems}
\label{subsec:causality}

Causality relies on a formal description of the interactions between observed variables in the form of a \textit{causal graph}.
Unlike black-box techniques, where predictions about an outcome are made with a pure data-driven approach, such graphical representation is effective when it comes to \textit{explainability}~\cite{causal-survey, causality-primer, bareinboim2022pearl}.
For this purpose, \oursystem generates a causal graph to: (i)~help domain experts understand the behavior of their IoT systems, (ii)~identify key metrics influencing the performance of the system, and (iii)~take more effective adaptation actions in response to changes.

\begin{definition} {\it (Causal graph).}
A causal graph $G$ (also called Causal Graphical Model) is the graphical representation that depicts the cause-effect relationships among variables in a given system.
$G =(\bf{V}, \bf{E})$ is structured as a Directed Acyclic Graph (DAG), where $\mathbf{V} = \{ \mu_{qos} \cup \mu_{context}\}$ are variables in the system\, and $\mathbf{E} \subseteq \mathbf{V} \times \mathbf{V}$ show causal effects (increase/decrease in latency, etc.).
$G$ represents a joint probability
distribution $P$ over a set of random variables $\mathbf{X} = (X_1, X_2, . . . , X_d)$ where $P$ is Markovian with respect to $G$.
\end{definition}
Through causal graphs, \oursystem is able to recognize the causes of QoS violations. This happens by identifying direct and indirect causes of QoS changes:
\begin{definition} {\it (Direct and indirect cause)}
For each directed edge $(X, Y) \in \textbf{E}$, $X$ is a \textit{direct cause} of $Y$ and $Y$ is a \textit{direct effect} of $X$. Recursively, every cause of $X$ that is not a direct cause of $Y$, is an \textit{indirect cause} of $Y$.
\end{definition}

\begin{figure}[htbp!]
  \centering
  \includegraphics[width=0.7\linewidth]{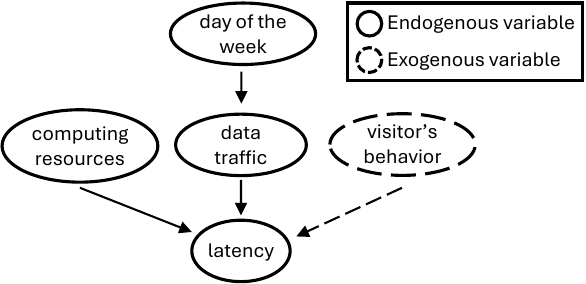}
  \caption{Causal Graphical Model}
  \Description{Sample causal graph.}
  \label{fig:causal-graph}
\end{figure}
Consider Figure~\ref{fig:causal-graph} with a sample causal graph that represents possible causal relationships in an IoT system. 
The graph shows that the \code{computing resources} and \code{data traffic} are a \textit{direct cause} of  \code{latency}.
In addition, since there is a \textit{direct path} from \code{day of the week} to \code{latency}, we define \code{day of the week} as an \textit{indirect cause} of latency.
We can further quantify the effects of causes on variables by building on the language of \textit{Structural Causal Models} to describe the collection of mechanisms underpinning a phenomenon.
\begin{definition}{\it (Structural Causal Model (SCM)).}
A structural causal model $\mathcal{M}$ is defined by a 4-tuple $\langle \textbf{V}, \textbf{U}, \mathcal{F}, \textit{P}(\textbf{U}) \rangle$:
\begin{itemize}[leftmargin=*]
    \item[--] $\textbf{V}$ is a set of \textit{endogenous} variables i.e., observable variables within the system that are determined by variables in the model. \textnormal{In IoT systems, these are the variables included in the set $\{\mu_{qos} \cup \mu_{context} \}$.}
    \item[--] $\textbf{U}$ is a set of \textit{exogenous} variables, i.e., unobservable variables where $V \cap U = \emptyset$. \textnormal{As shown in Figure~\ref{fig:causal-graph}, even though \code{visitor's behavior} (e.g., preferences regarding exhibition halls) directly affects latency, \oursystem would not observe this variable since it can neither be controlled nor monitored by the system. Such variables  cannot be found in $\{ \mu_{qos} \cup \mu_{context}\}$.}
    \item[--] $\mathcal{F}$ is a set of functions ${f_i, ..., f_n}$, where each $f_i$ $\in$ $\mathcal{F}$ is defined as $f_i: (\textbf{V} \cap \textbf{U})^p \rightarrow \textbf{V}$, with $p$ the ariety of $f_i$ so that $f$ determines completely the value of $V_i$. $f_{i}$ enables quantifying the effect of changing a variable in $G$ on its child. \textnormal{For instance, a $2 \times$ increase in \code{network traffic} may result in a $1.5 \times$ increase in \code{latency}, i.e., $f(\code{latency}) = 1.5 \times \code{network traffic}$.}
    \item[--] $P(\textbf{U})$ is the joint probability distribution over the exogenous variables $P(\textbf{U}) = \prod_{i} P(U_i)$.
\end{itemize}
\end{definition}

In most application domains, the causal graph may only be partially known by experts. 
To better understand a system behavior, \oursystem recovers the complete causal graph.

\begin{definition}{\it (The causal discovery problem).}
The causal discovery problem consists in recovering the \textit{true} graph $G^*$ from a given dataset $\textbf{D}$.
\end{definition}
This requires \oursystem to find $\textbf{V} = \{ \mu_{qos} \cup \mu_{context}\}$ by relying on a causal discovery algorithm.

\begin{definition}{\it (Causal Discovery Algorithm).}
 A causal discovery algorithm is said to \textit{solve} the causal discovery problem iff. it converges to the true graph $G^*$ in the limit of the sample size of the dataset $\textbf{D}$. 
\end{definition}
Such algorithms aim to identify the underlying DAG representing the causal influences in a system. 
Popular examples include the Peter and Clark (PC)~\cite{spirtes2001causation}, Greedy Equivalence Search (GES) \cite{chickering2002optimal}, and the Fast Causal Inference (FCI)~\cite{spirtes2001causation} algorithms.
Note that causal discovery algorithms are typically non-deterministic, and therefore it is necessary to validate the generated causal graph with domain experts to validate their correctness. The dataset quality and richness affects the graph quality, and therefore it is better to include as many relevant variables as possible when collecting the dataset.
\newline
In some systems, a cause-effect pair is known to exist (or to not exist) \textit{a priori}, e.g. through domain expert’s elicitation.
With \oursystem, domain experts can explicitly represent such pairs via a knowledge base used by the causal discovery algorithm to inform and constrain the structure of the causal graph. 
This can be achieved either manually by domain experts, or by using existing Ontologies that represent the IoT system.
\begin{definition}{\it (Knowledge base).}
A knowledge base $\mathcal{K}$ is defined as an ordered pair $(\textbf{R}, \textbf{F})$ where $\textbf{R}$ is the set of required directed edges, while $\textbf{F}$ is the set of forbidden directed edges.
\end{definition}

\subsection{CRL-based Adaptation}
\label{subsec:rl}
\oursystem relies on Reinforcement Learning~\cite{rl} as a decision-making approach for taking adaptation actions for satisfying the QoS requirements of applications in IoT environments.
However, unlike traditional approaches where states and actions are chosen regardless of causal relationships, \oursystem enhances decision-making abilities by using the insights gained from causal graphs.
The generated causal graph is used to define the set of states and actions that the \oursystem adaptation agent can act upon.
This process can be achieved either manually, or by using automated approaches to convert nodes in the causal graph to states and actions.
We provide next a formal definition of the RL problem this paper deals with.

Reinforcement Learning requires an agent to interact with an environment over time to learn the best action(s) to take under specific situations, to optimize an objective function (e.g., minimizing latency).
At each time step $t$, the agent receives a state $s_t$ in a state space $\mathcal{S}$ and selects an action $\alpha_t$ from an actions space $\mathcal{A}$, following a policy $\pi(\alpha_t \mid s_t)$.
The policy represents the agent's behavior, i.e., a mapping from $s_t$ to $\alpha_t$.
When an action is taken, the agent receives a scalar reward $r_t$ according to a reward function $\mathcal{R}(s, \alpha)$, and transitions to the next state $s_{t+1}$ following a state transition probability $\mathcal{P}(s_{t+1} \mid s_t, \alpha_t)$.
For instance, an agent may decide to allocate more bandwidth to network switches in high-traffic areas (action $\alpha_t$) to reduce data exchange latency (state $s_t$). The reward $r_t$ can be based on how effectively the bandwidth allocation reduces latency while maintaining overall network efficiency. After this action, the system transitions to the next state $s_{t+1}$, where new metrics such as updated latency and network load are observed.
The state transitions and rewards are stochastic and are assumed to have the Markov property.
In an episodic problem, this process continues until the agent reaches a terminal state and then it restarts.
This can be a state where QoS requirements of all IoT application are satisfied.
The return $R_t = \sum_{k=0}^{\inf} \gamma^k r_{t+k}$ is the discounted, accumulated reward with the discount factor $\gamma \in (0, 1]$.
The agent aims to maximize the expectation of such long term return from each state, which leads to learning the optimal policy $\pi^\ast$ to follow. 

In existing approaches, the state space $\mathcal{S}$ is usually modelled based on $\vSystemProperties$, which may include network link congestion, data traffic, and resource consumption~\cite{adaptation-survey}.
However, \oursystem identifies additional relevant variables (including $\vContextProperties$) that affect system performance through Causal Discovery, as discussed in Section~\ref{subsec:causality}.
Hence, based on the obtained causal model, \oursystem defines the RL problem as follows:
\begin{itemize}[leftmargin=*]
    \item [--] The set of states $\mathcal{S}$ is defined using $\vContextProperties$ (e.g., location, number of occupants, day and time).
    \item [--] The set of actions $\mathcal{A}$ includes the available actions that can be taken for adaptation (e.g., bandwidth/resource allocation).
    \item [--] To satisfy the QoS requirements of applications, \oursystem defines a reward function $\mathcal{R}$ to minimize the deviation between the QoS requirements and the observed performance of the system.
    \begin{equation}
    \label{eq:reward-function}
        \mathcal{R} = - (\sum_{\vApplication \in \vApplications} \phi_{i}  \frac{\vLatency_{\vApplication} - \vLatency_{qos}}{\vLatency_{qos}} + c_1 . \omega + c_2 . \vResources(\vEdgeNode))
    \end{equation}
    
where $\phi_i$ represents the ``priority'' of the application (i.e., how important the application is), and $\frac{\vLatency_{\vApplication} - \vLatency_{qos}}{\vLatency_{qos}}$ represents the deviation between the latency defined by the application's QoS requirements and the actual latency perceived by the application.
To avoid over-allocation of resources, we add penalty terms ($c_1$ and $c_2$) to the bandwidth and resources allocated in the system.
\end{itemize}

Two main approaches exist for solving RL problems.
Value-based methods --- such as Q-learning/Deep Q-learning --- focus on learning an optimal value function by mapping states and actions to rewards~\cite{rl}. 
However, these methods are more suited for environments where states and actions can be modelled as discrete values.
In more complex environments with a continuous action space (e.g., controlling resource allocation in complex IoT systems), such methods may struggle to find the best policy~\cite{rl}.
For this purpose, \oursystem relies on policy-based methods, such as Proximal Policy Optimization (PPO)~\cite{ppo}, to directly learn the optimal policy $\pi^\ast$ without having to learn a value function.

\section{Experimental Evaluation}
\label{chap:crafter/sec:experiments}
This section presents the evaluation of \oursystem using a realistic smart museum scenario.
For this purpose, we first implement the \oursystem prototype for simulating IoT data exchange in smart spaces (Section~\ref{subsec:prototype-implementation}).
This enables modeling spaces and defining the deployed devices, virtual sensors, and applications.
In addition, the prototype supports controlling the networking infrastructure, resource allocation parameters, and number of occupants in each location of the smart space.
Then, we model several exhibition halls in the Louvre by relying on the analysis done in ~\cite{louvre-paper}, which monitors visitors' movement patterns throughout various locations.
Using the developed prototype and models, we emulate IoT data exchange and generate an IoT metrics dataset containing response times per data flow (Section~\ref{subsec:exp-setup}).
\oursystem then extracts causal relationships between the different variables in the smart museum scenario (Section~\ref{subsec:exp-causal-discovery}).
To showcase \oursystem's decision-making capabilities, we feed the smart museum causal graph to \oursystem's Adaptation Agent for taking adaptation decisions to keep the response time of applications under the QoS requirements that they define (Section~\ref{subsec:exp-rl}).
Finally, we compare our solution against state-of-the-art approaches (Section~\ref{subsec:evaluation}).
The prototype developed, the models and datasets used for the following experiments, along with the solution and results are publicly available on  
\url{https://github.com/satrai-lab/crafter/}.

\subsection{\oursystem Prototype Implementation}
\label{subsec:prototype-implementation}

\begin{figure}[t]
  \centering
  \includegraphics[width=\linewidth]{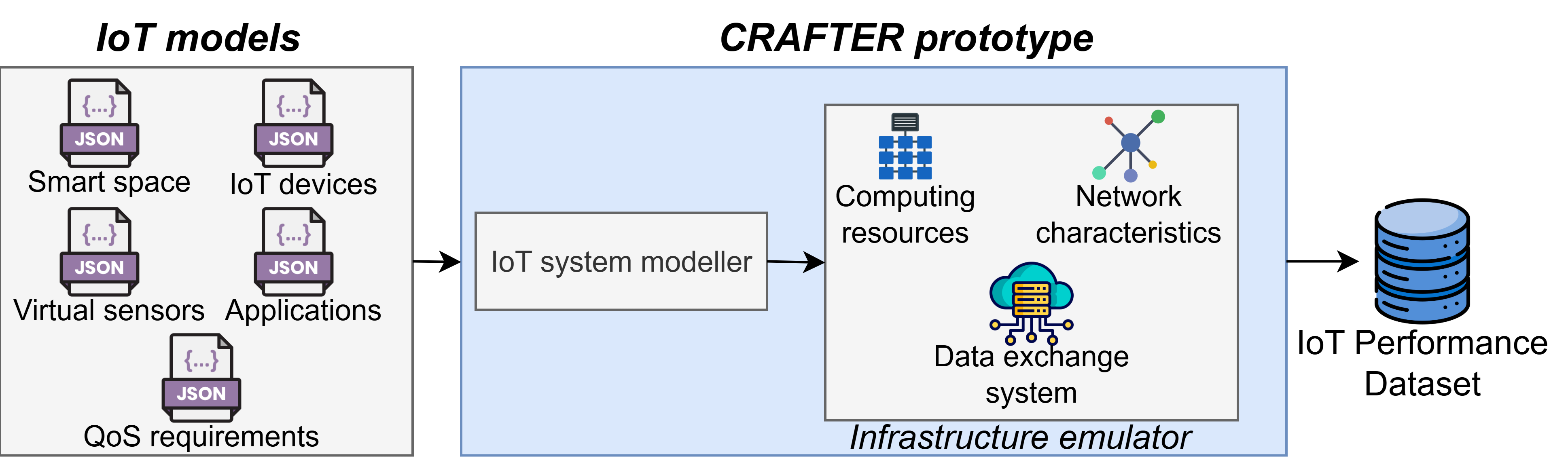}
  \vspace{-0.3in}
  \caption{The \oursystem prototype implementation}
\vspace{-0.05in}
\Description{Figure showing the CRAFTER prototype implementation.}
  \label{fig:prototype-implementation}
\end{figure}

\begin{table}[]
\caption{Device and visitor allocation per location}
\vspace{-0.1in}
\label{tab:device-allocation}
\Description{Table showing device and visitor allocation per location.}
\begin{tabular}{|c|c|c|}
\hline
Location & \# devices & \% visitors \\ \hline
B        & 25        & 35         \\ \hline
C        & 11        & 5          \\ \hline
D        & 17        & 5          \\ \hline
E        & 18        & 15         \\ \hline
G        & 16        & 15         \\ \hline
S        & 15        & 10         \\ \hline
V        & 10        & 15         \\ \hline
\end{tabular}
\end{table}

\begin{table*}[]
\caption{Properties of virtual sensors and applications}
\label{tab:app-properties}
\Description{Table including properties of virtual sensors and applications used in the experimental setup.}
\small
{\renewcommand{\arraystretch}{1.3}
\scalebox{0.81}{
\begin{tabular}{|c|c|c|c|c|c|c|}
\hline
\textbf{Virtual Sensor}               & \textbf{Received data}                                                                         & \textbf{Processing technique}                                                                                              & \multirow{8}{*}{\textbf{}} & \textbf{Application}             & \textbf{Received data}                                & \textbf{QoS requirements} \\ \cline{1-3} \cline{5-7} 
\multirow{2}{*}{Occupancy Prediction} & \multirow{2}{*}{\begin{tabular}[c]{@{}c@{}}Camera feed, \\ visitor's location\end{tabular}}    & \multirow{2}{*}{\begin{tabular}[c]{@{}c@{}}CNN for people detection, \\ aggregation with visitors' locations\end{tabular}} &                            & \multirow{2}{*}{Metaverse}       & \multirow{2}{*}{Lighting, noise, visitors' locations} & \multirow{2}{*}{20 ms}    \\
                                      &                                                                                                &                                                                                                                            &                            &                                  &                                                       &                           \\ \cline{1-3} \cline{5-7} 
\multirow{2}{*}{Queue Management}     & \multirow{2}{*}{Visitor's location}                                                            & \multirow{2}{*}{\begin{tabular}[c]{@{}c@{}}Estimation of waiting times \\ using linear regression\end{tabular}}            &                            & Smart lighting                   & Lighting                                              & 100 ms                    \\ \cline{5-7} 
                                      &                                                                                                &                                                                                                                            &                            & \multirow{2}{*}{Visitor guiding} & \multirow{2}{*}{Occupancy, queueing times}            & \multirow{2}{*}{400 ms}   \\ \cline{1-3}
\multirow{2}{*}{Security Analysis}    & \multirow{2}{*}{\begin{tabular}[c]{@{}c@{}}Camera feeds, motion ,\\ door sensors\end{tabular}} & \multirow{2}{*}{\begin{tabular}[c]{@{}c@{}}Aggregation of data for security\\ threat detection\end{tabular}}               &                            &                                  &                                                       &                           \\ \cline{5-7} 
                                      &                                                                                                &                                                                                                                            &                            & Artifact maintenance             & Maintenance date                                      & 2 s                       \\ \cline{1-3} \cline{5-7} 
Predictive Maintenance                & \begin{tabular}[c]{@{}c@{}}Temperature, humidity, air quality, \\ proximity\end{tabular}       & \begin{tabular}[c]{@{}c@{}}Estimation of maintenance date\\ using linear regression\end{tabular}                           &                            & Security                         & Security alert                                        & 1s                        \\ \hline
\end{tabular}}}
\end{table*}

For evaluating \oursystem, we implement a Java-based prototype for emulating IoT data exchange in smart spaces.
As shown in Figure~\ref{fig:prototype-implementation}, we model properties of smart spaces, devices, virtual sensors, and applications and their QoS requirements in JSON format.
We consider that a space is divided into multiple physical locations that may have different characteristics.
These include the area of the space, the number of occupants in the space, and the type and number of IoT devices deployed.
Then, we model devices' properties: their type (e.g., temperature sensor, motion sensor), the observation(s) they provide (e.g., temperature, humidity, number of occupants), and the distribution and rate at which they generate data.
Virtual sensors define observations that they receive and process using simple rule-based algorithms or more complex ML-based approaches.
Finally, applications deployed in the space specify observations to be received from devices and virtual sensors, which are then translated to topic subscriptions.
They also define QoS requirements related to latency, throughput, energy consumption, and other metrics that have to be satisfied.
Communication between the different components of the IoT system takes place through a publish/subscribe architecture.
For this purpose, we use the MQTT Paho Library~\cite{mqtt-paho} for implementing MQTT clients (devices, virtual sensors, and applications), and EMQX 5.6.1~\cite{emqx} as a message broker for handling topic-based data exchange.
To emulate the networking infrastructure, we rely on Containernet~\cite{containernet}, a fork of the Mininet network emulator that supports using Docker containers as network hosts.
Each location is packaged and deployed as a Docker container (i.e., network host) connected to its dedicated switch; this allows controlling the characteristics of each location's network link separately.
In addition, Edge and Cloud resources are emulated to host virtual sensors and heavyweight computing tasks.

\begin{figure}[t]
  \centering
  \includegraphics[width=\linewidth]{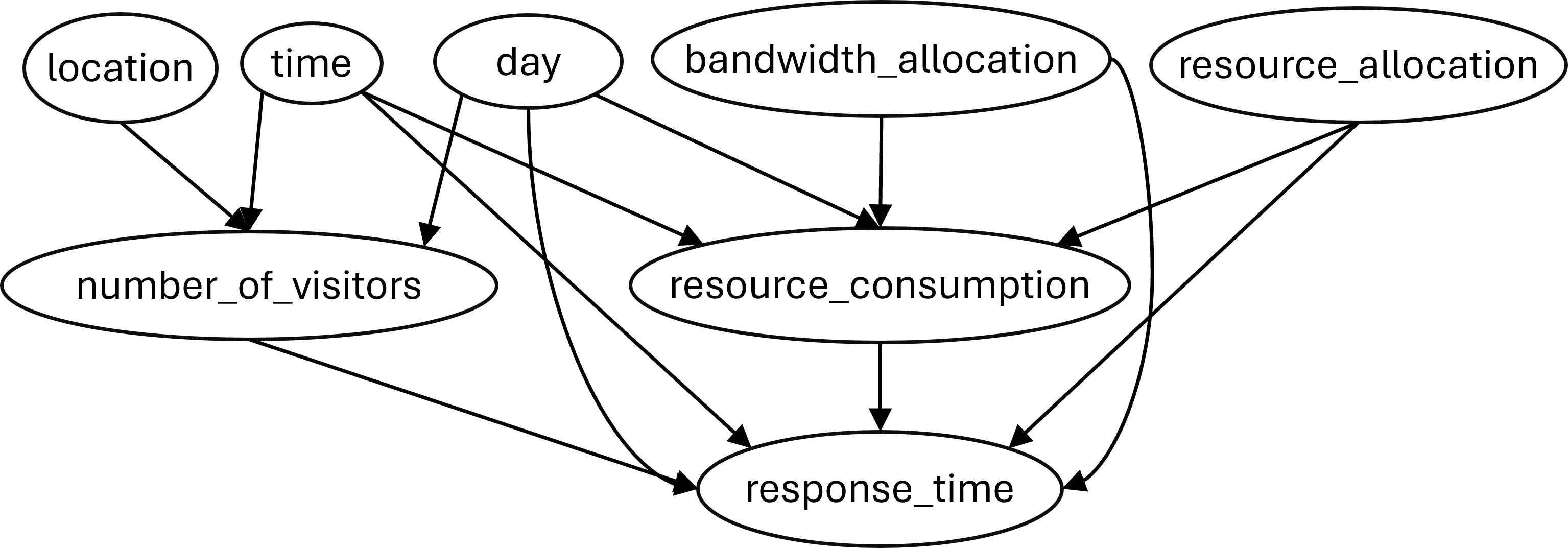}
  \caption{Causal graph for the smart museum scenario}
  \label{fig:exp-causal-graph}
  \Description{CRAFTER-generated causal graph for the smart museum scenario.}
\end{figure}

\subsection{Experimental Setup}
\label{subsec:exp-setup}
To evaluate our approach, we rely on the analysis done in~\cite{louvre-paper} to model 7 exhibition halls in the Louvre, containing popular paintings and artifacts such as Da Vinci's \textit{Mona Lisa} and Canova's \textit{Psyche and Cupid}, and to distribute visitors across locations.
We consider that each exhibition hall contains IoT devices deployed for monitoring various space properties; in total, $112$ devices are deployed across the smart space.
Table~\ref{tab:device-allocation} shows the allocation of the number of devices and visitors per location, and Table~\ref{tab:app-properties} shows the properties of the virtual sensors and applications deployed.
We consider that each visitor carries a mobile device that connects to the Louvre infrastructure for receiving visitor guiding information, and periodically sends its location information to help Louvre administrators' analyze visitor's movements and the most visited artifacts.
Two Edge nodes are available to host 4 types of virtual sensors.
Finally, we consider 5 representative application types that are deployed in the museum, and that define latency QoS requirements to be met.
As described in Section~\ref{subsec:louvre-scenario}, the Louvre sub-metaverse application enables remote visitors to visit the Louvre in real-time via VR equipment.

\subsection{Extracting Causal Models}
\label{subsec:exp-causal-discovery}

As described in Section~\ref{subsec:overview}, at design-time, \oursystem extracts causal relationships in the IoT system.
Recall from Section~\ref{subsec:causality} that generating causal models involves collecting an IoT metrics dataset $\mathbf{D}$.
For this purpose, we use the \oursystem prototype to emulate IoT data exchange in the Louvre setup presented in Section~\ref{subsec:exp-setup}.
To generate the dataset $\mathbf{D}$, we emulate data exchange in the Louvre for each day of the week (excluding Tuesday, when the Louvre is closed) and for two time periods each day: the morning period (AM) and evening period (PM).
We vary the number of visitors based on the day and time period, with the weekend days (Friday, Saturday, Sunday) being the most crowded  of the week~\cite{louvre-paper}.
For each day and time period, we vary the bandwidth value from each location from $20$~Mbps to $500$~Mbps, and vary the percentage of resources allocated to virtual sensors from $10\%$ of total resources to $100\%$.
In total, we generate a dataset that captures $1560$ different possible configurations of the system.
For each configuration, we record the response time for individual messages received by each application, as well as the resource consumption per Edge node.

Once the dataset is generated, we use \oursystem to extract causal models.
We use the Causal-Learn Python library~\cite{zheng2024causal} for generating the causal graph in \code{DOT} format from the IoT dataset $\mathbf{D}$.
For this purpose, we rely on the Peter-Clark (PC) \cite{spirtes2001causation, Le2015AFP} algorithm using Kernel-based Conditional Independence~\cite{zhang2012kernel} for testing for independent variables.
PC is known for its ability to easily identify conditional independencies, and offers scalability and flexibility in handling various kinds of data. 
PC starts with a complete graph where each node represents a variable in $\mathbf{V}$, and each edge represents a causal relationship.
The algorithm then iteratively tests for conditional independencies and removes edges between nodes that are conditionally independent given any subset of the other nodes.
Rules such as orientation based on V-structures and propagation of direction~\cite{spirtes2001causation} are then applied to orient edges following the existing independencies between variables and the provided knowledge base.
The generated causal graph for the smart museum setup is shown in Figure~\ref{fig:exp-causal-graph}.
We can see that bandwidth allocation, resource allocation, and resource consumption are a direct cause of response time.
Naturally, changing these parameters usually results in better/worse response time values for messages.
However, we can also observe the context variables that have direct and indirect effects on the response time experienced by applications.
For instance, the day and time are direct causes of response time, since they are also causes of fluctuations of the number of visitors.
Location, on the other hand, is an indirect cause of response time, since visitors' distribution across exhibition halls in the museum varies based on the popularity of artifacts and paintings.
Hence, when more visitors are present in a hall, data traffic will increase in that location, leading to more congested network links, and a need for more computing power for virtual sensors to process the generated data in a timely manner.
We define $\mu_{context} = \{$\code{day of the week, time of the day, number of visitors}$\}$, and $\mu_{qos} = \{$\code{bandwidth, resource allocation, resource consumption}$\}$.

The causal graph enables more formal explanation of the performance of IoT systems, without the need for assumptions and trial-and-error testing.
In more complex deployments, such causal models are not trivial to retrieve only through expert elicitation, hence the need for causal discovery approaches.

To highlight \oursystem's ability to generate causal graphs across different IoT domains with minimal domain expertise, we evaluate \oursystem on three different real-world datasets spanning different IoT use cases.
These include: (i)~an Autonomous Moving Robot (AMR) scenario with network throughput and latency requirements, (ii) a smart factory scenario involving a copper production manufacturing plant with energy consumption requirements, and (iii)~an anomaly detection scenario in a production cell with latency and throughput requirements.
Our evaluation shows that \oursystem manages to generate relevant causal graphs for three use cases, illustrating how the same self-adaptation can be applied to different IoT domains.
The generated causal graphs and more information regarding the evaluation scenarios can be found on: \url{https://github.com/satrai-lab/crafter/}.

\subsection{Learning Self-adaptation with Causal RL}
\label{subsec:exp-rl}

\begin{figure}[t]
  \centering
  \includegraphics[width=\linewidth]{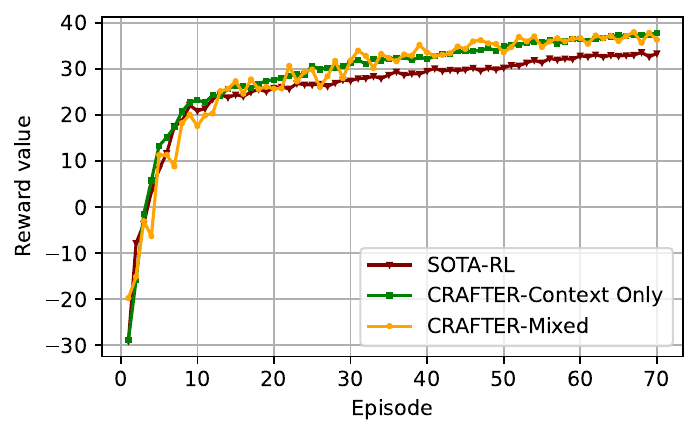}
  \caption{Reward per training episode}
\vspace{-0.2in}
  \label{fig:training}
  \Description{Graph showing RL reward per training episode. CRAFTER-Mixed achieves the higher reward per training episode.}
\end{figure}

\begin{figure}[t]
  \centering
  \includegraphics[width=\linewidth]{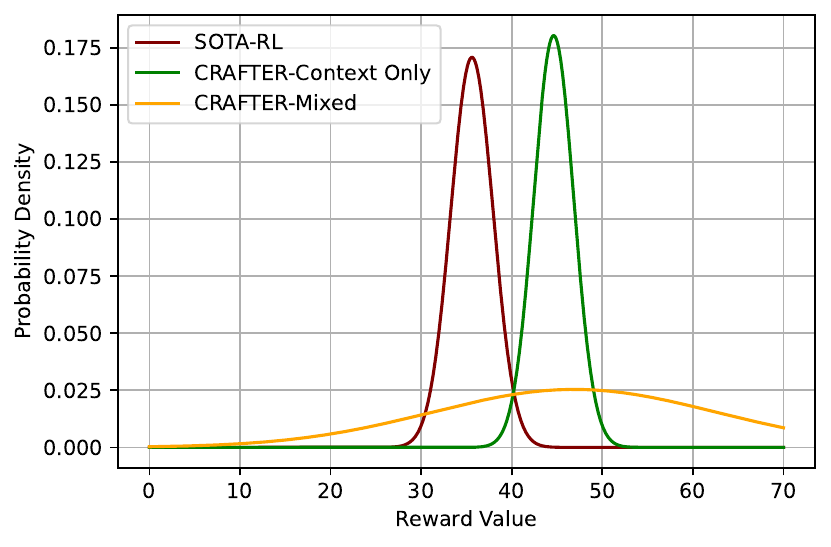}
  \caption{Distribution of evaluation rewards}
\vspace{-0.2in}
  \label{fig:evaluation}
  \Description{Figure showing the distribution of evaluation rewards. CRAFTER-Context Only and CRAFTER-Mixed achieve the highest reward value.}
\end{figure}

\begin{figure*}[t]
  \centering
  \begin{subfigure}[t]{0.32\textwidth}
    \centering
    \includegraphics[width=0.95\textwidth]{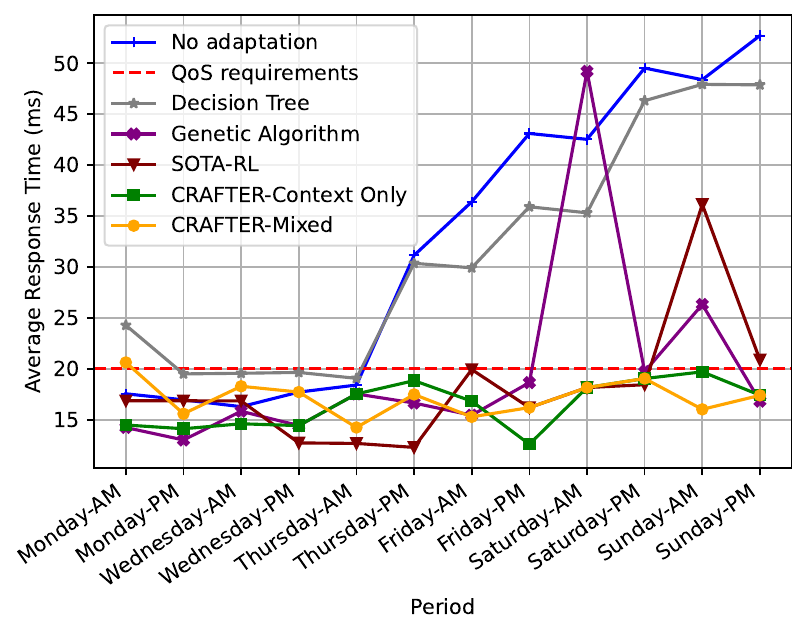}
    \caption{Metaverse application}
    \label{fig:subfig1}
  \end{subfigure}
  \hfill
  \begin{subfigure}[t]{0.32\textwidth}
    \centering
    \includegraphics[width=\textwidth]{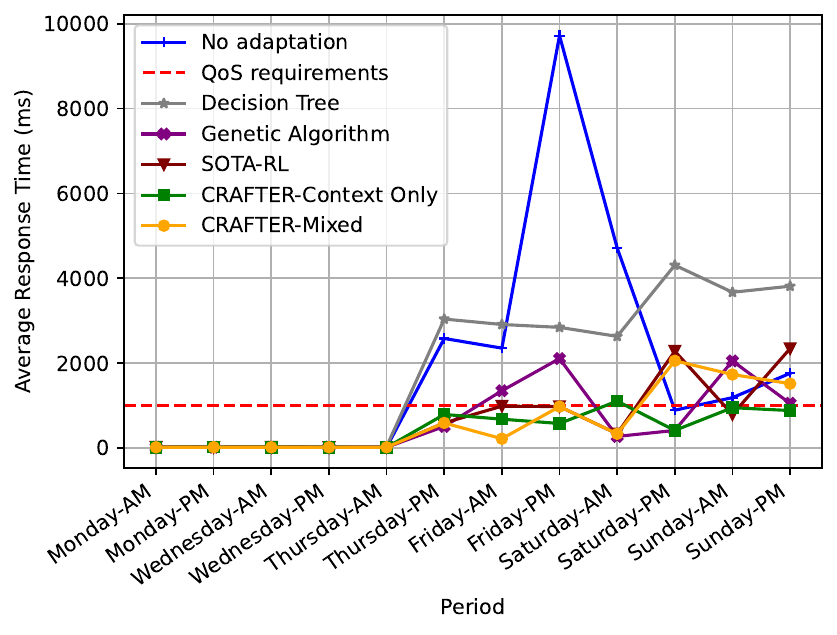}
    \caption{Security application}
    \label{fig:subfig2}
  \end{subfigure}
  \hfill
  \begin{subfigure}[t]{0.32\textwidth}
    \centering
    \includegraphics[width=\textwidth]{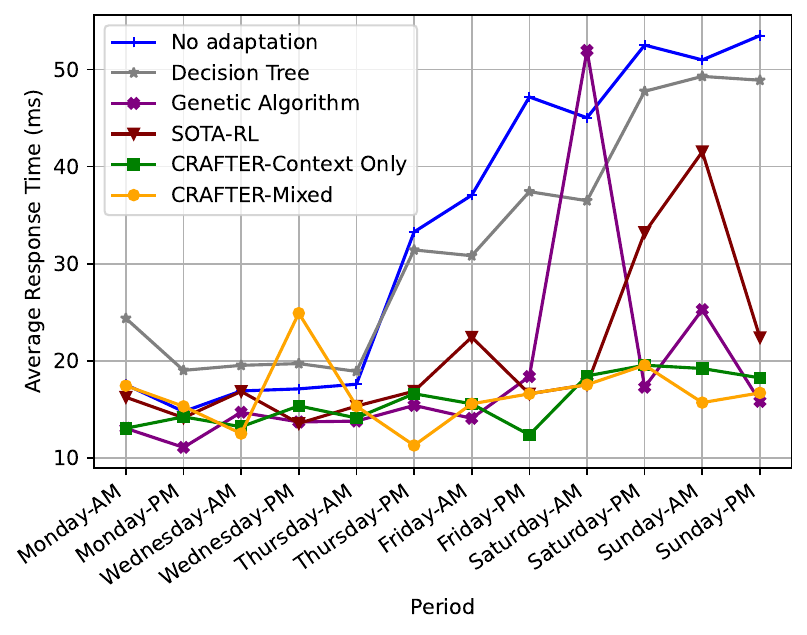}
    \caption{Smart lighting application}
    \label{fig:subfig3}
  \end{subfigure}
  
   \hspace{0.5in}
  \begin{subfigure}[t]{0.32\textwidth}
    \centering
    \includegraphics[width=\textwidth]{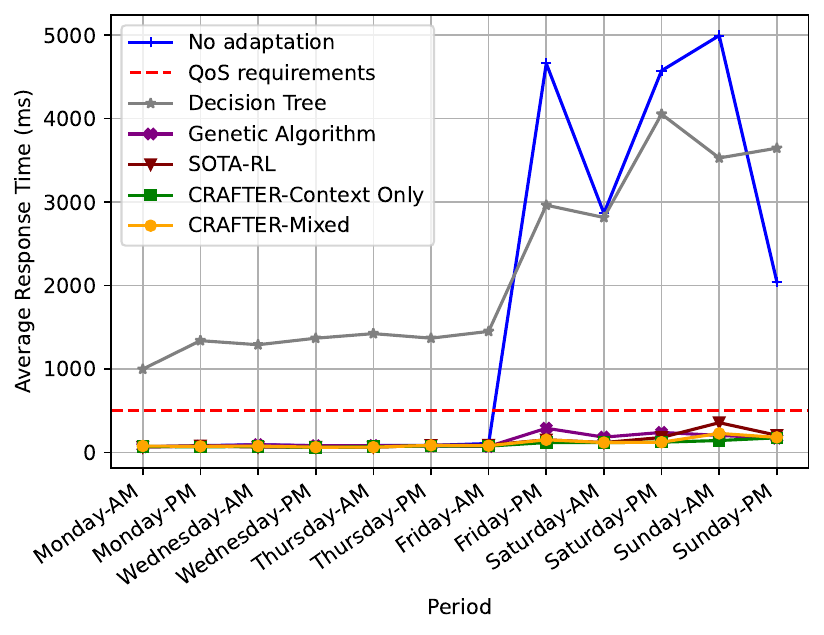}
    \caption{Visitor guiding application}
    \label{fig:subfig4}
  \end{subfigure}
  \hspace{0.5in}
  \begin{subfigure}[t]{0.32\textwidth}
    \centering
    \includegraphics[width=\textwidth]{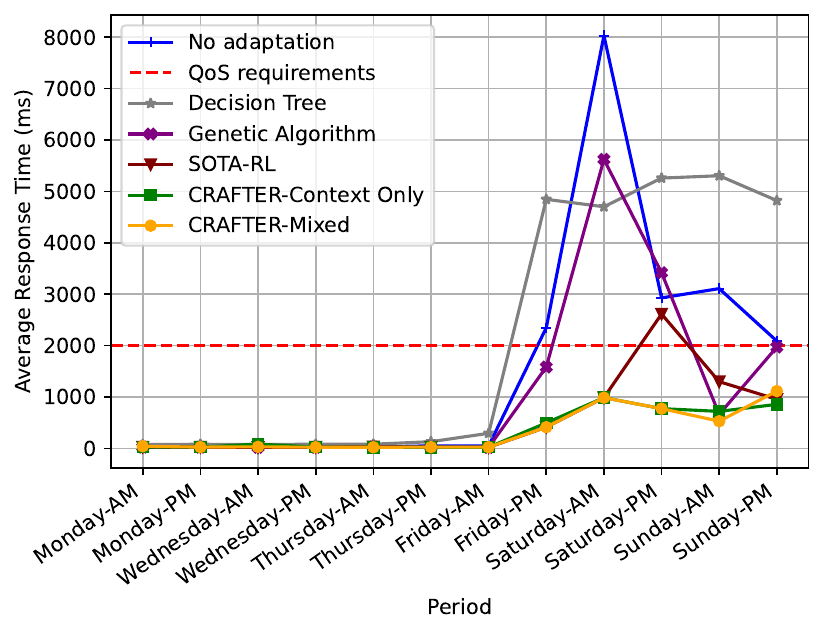}
    \caption{Maintenance application}
    \label{fig:subfig5}
  \end{subfigure}
  \caption{Average response time per application}
  \label{fig:response-times}
  \Description{Graphs showing response time per application of CRAFTER vs. existing approaches.}
\end{figure*}

\oursystem's Adaptation Agent is guided by the causal graphs generated in Section~\ref{subsec:exp-causal-discovery} for taking efficient adaptation actions.
We train two \oursystem Adaptation Agents using the obtained causal graph: a first agent (called \code{\oursystem-Context Only}) that observes states based only on the $\vContextProperties$ identified by the causal model.
The second agent (called \code{\oursystem-Mixed}) observes both $\vContextProperties$ and $\vSystemProperties$.
The possible actions to be taken include allocating a specific bandwidth to network links for the different locations, and allocating computing resources to Edge nodes where virtual sensors are deployed.
Training the RL agents involves interacting with the IoT environment to learn the optimal policy.
When an agent selects an action to be performed, the \oursystem prototype is simulated under the configuration chosen by the agent, and the reward function value and the next state values are returned to the agent through a feedback loop.
We use the Stable-Baseline3 library~\cite{stable-baselines3} to design the RL agents, and Gymnasium~\cite{towers_gymnasium_2023} for creating the training environment.
We train the agents using the PPO algorithm~\cite{ppo} to optimize the reward function defined in Equation~\ref{eq:reward-function}, which aims to reduce each application's response time deviation w.r.t. its QoS requirements, while avoiding over-allocation of network and computing resources.
The agents are trained for $70$ episodes using a learning rate of $0.0003$ and a clip range of $0.2$.

\subsection{\oursystem Against Existing Approaches}
\label{subsec:evaluation}
We evaluate \oursystem against (i)~state-of-the-art RL-based approaches commonly used for adaptation in IoT systems~\cite{ream, chen2021deep, 9555233}, (ii)~approaches that use genetic algorithms for adaptation~\cite{genadapt}, and (iii)~rule-based approaches~\cite{aquacep, planiot}.

We first evaluate the training performance of RL agents.
In addition to \code{\oursystem-Context Only} and \code{\oursystem-Mixed}, we train an RL agent to take adaptation decisions based solely on $\vSystemProperties$ (called \code{SOTA-RL}).
\code{SOTA-RL} is trained in the same environment as \oursystem agents, using the same hyperparameters and for the same number of episodes.
For all three agents, the training algorithm, the action space and reward function (Eq.~\ref{eq:reward-function}) are the same. 
Figure~\ref{fig:training} shows the reward value per training episode for each of the agents.
\texttt{\oursystem-Context Only} achieves a $22\%$ improvement in training compared to \texttt{SOTA-RL}.
This is due to the incorporation of key context variables in the observed states identified through the causal models.
By exploiting valuable cause-effect relationships, RL agents are able to make better and more informed adaptation decisions.
The improvement is even more pronounced when we combine $\vSystemProperties$ and $\vContextProperties$: the \texttt{\oursystem-Mixed} approach achieves up to $30\%$ improvement in training over \code{SOTA-RL}, albeit with a lower training stability than \texttt{\oursystem-Context Only}.
We next evaluate the three models over 100 episodes, and record the mean and standard deviation of the obtained rewards for each episode.
Figure~\ref{fig:evaluation} shows the distribution of rewards during the evaluation episodes.
\texttt{\oursystem-Context Only} once again achieves a better performance than \texttt{SOTA-RL}, with a $25\%$ better mean reward value.
The standard deviation of reward distribution is also smaller for \texttt{\oursystem-Context Only} ($2.21$) than for the \texttt{SOTA-RL} approach ($2.33$).

Next, we evaluate the adaptation capabilities of \oursystem using the setup presented in Section~\ref{subsec:exp-setup}.
To compare with genetic programming approaches, we use the \textit{NSGA-II}~ algorithm~\cite{nsgaii} implemented with the \textit{DEAP}~\cite{deap} Python library, with a population size of $50$, $70$ generations, a mutation probability of $0.2$ and a crossover probability of $0.5$.
To compare with rule-based approaches, we train a Decision Tree implemented with \textit{Scikit-learn}~\cite{scikit-learn}.
For each application type, we plot the average response time per day and time period under the following conditions: (i)~without adaptation, i.e., the same configuration (bandwidth and resource allocation) is used throughout the whole week, (ii)~with adaptation using the state-of-the-art approaches, (iii)~using the \texttt{\oursystem-Context Only} and \texttt{\oursystem-Mixed} adaptation approaches.

Figure~\ref{fig:response-times} shows the average response time per application.
As expected, without any adaptation solution, the QoS requirements of most applications are violated when the properties of the smart environment change.
For example, as the number of visitors and devices increases in the museum starting from Thursday afternoon, the response time of applications increases significantly for all applications.
The Decision Tree approach provides better results than static configurations, but its rule-based structure limits its ability to cope with more dynamic and complex fluctuations in traffic and resource demand. This results in high response times during peak conditions (with a response time of up to $47~ms$ for Metaverse applications).
This shows that simply using rule-based approaches for adaptation purposes is not enough for satisfying the QoS requirements of applications in highly dynamic IoT environments.
The Genetic Algorithm, on the other hand, enables more effective adaptation actions as the number of visitors and the workload vary.
For example, it manages to keep the latency of visitor guiding applications ($478~ms$) under the QoS requirements of $500~ms$. 
However, it struggles to maintain the required QoS consistently for other applications, especially under a high workload.
On weekend days, the latency for Metaverse applications reaches up to $49~ms$ during peak visiting hours.
While the Genetic Algorithm can adapt to some degree, it fails to keep response times under strict thresholds for more demanding applications.
In response to changes, \texttt{\oursystem-Context Only} manages to keep the response time for all applications under their defined QoS requirements, even when the number of visitors reaches a peak, on Saturday and Sunday, as shown in Figure~\ref{fig:response-times}.
Even for Metaverse applications that require an ultra-low latency of less than $20~ms$, \oursystem effectively adapts to the changing properties of the IoT system.
The \texttt{\oursystem-Mixed} approach also manages to offer appropriate adaptation in dynamic situations, keeping the response time of Metaverse under $20~ms$ as well.
On the other hand, the \texttt{SOTA-RL} approach fails to keep the response time of messages under $20~ms$, resulting in a response time of over $35~ms$ ($75\%$ higher than the QoS requirements of Metaverse applications), especially during the busiest periods.
The same trend is observed for other applications, where \texttt{\oursystem-Context Only} succeeds in keeping response time values of applications under the specified requirements.
By exploiting relevant context information that have an effect of the performance of the system, \oursystem can effectively adapt to changes in dynamic environments.

\section{Related Work}
\label{chap:crafter/sec:related}
This section provides an overview of literature related to self-adaptive IoT systems and causal discovery.

\textbf{Self-adaptive IoT.} Significant research has been carried out on self-adaptive systems recently.
Application layer approaches such as  \cite{aquacep}, \cite{10.1145/3365871.3365874} focus on  dynamically adapting sensor deployment and data source assignment using quality evaluation and opportunistic sensing to meet application requirements and system changes.
Distream~\cite{distream} introduces dynamic workload balancing for low-latency, high-throughput live video analytics. However, it only focuses on task distribution for video streaming application.
\cite{9101225} proposes a goal-driven architecture for adapting to sudden device unavailability and changing deployment topologies, while~\cite{10.1145/3410992.3411003} employs a machine learning for user activity recognition for the deployment of goal-driven IoT systems.
\cite{planiot, planemqx} leverage AI planning for adaptive data flow configuration in IoT environments based on QoS requirements and dynamic application changes.
Middleware approaches such as Sora~\cite{sora} introduce a Scatter-Concurrency-Goodput method that focuses on adapting soft resources to critical microservices,
while~\cite{mendula2024furcifer} offers a framework for adaptation of cloud computing continuum configurations in dynamic mobile settings.
Similarly, \cite{pandora-paper} provides a framework for ensuring autonomous AIoT operation in complex deployments spanning the computing continuum. 

RL-based methods have become prominently used for self-adaptive IoT systems~\cite{chen2021deep}.
For instance, the authors of~\cite{9239654} present an architecture using LSTM networks for forecasting data traffic and RL for adaptation to enhance energy efficiency in IoT systems.
REAM~\cite{ream} provides a publish/subscribe middleware for resource monitoring based on application priority and employs RL agents for monitoring and adaptation actions.
\cite{9555233} uses Deep RL for load-balancing computing tasks in IoT Edge systems while taking into account the communication requirements of applications.
\cite{zhang2024safety} introduces safety-aware adaptive RL for autonomous mobile assistive robots, where the agent must adapt actions at runtime while respecting safety constraints and environmental changes. The approach is designed for real-world robotic agents operating in dynamic environments typical of self-adaptive systems. While this approaches emphasizes runtime safety constraints and robotics adaptation, \oursystem targets IoT systems and couples RL with design-time causal models to reduce exploration overhead and identify causal drivers of performance issues.
\cite{9659519} proposes a meta-reinforcement learning approach for self-adaptive systems that can quickly adapt policies across multiple possible environment dynamics. It learns a meta-policy across models so that when the real environment appears, the system can adapt faster. On the other hand, \oursystem uses causal discovery and RL in a unified framework to try and anticipate root causes of performance issues, and not just quickly re-learn policies, but understand what drives adaptation needs.
\cite{scotti2025antifragility} uses online learning and monitoring techniques in an IoT context to achieve antifragility — systems that can improve performance under stress through continuous learning.
\cite{edrisi2025approaching} proposes a proactive approach for the self-adaptation of non-linear cyber-physical systems based on Adaptive Model Predictive Control (AMPC), and is designed primarily for autonomous moving robots.

While the above approaches support self-adaptation when system performance is degraded, adaptation actions are reactive (i.e., after the QoS violations are detected). They also fail to identify the causes of QoS violations.
\oursystem supports proactive adaptation by understanding the variables that impact IoT system performance.

\textbf{Causal discovery.}
Causality has been traditionally used in various fields, such as genomics, social sciences, economics, among others~\cite{causal-survey}.
However, with the growing need for explainable systems, causality-based approaches are being more widely used in networking and AI-based approaches.
For example, \cite{Wang2024ACT} proposes wireless traffic prediction using Granger Causal Discovery and a GNN-based model to explore spatio-temporal causal relationships between network variables. \cite{Gui2023CaFANetCA} introduces CaFANet that uses causal analysis and attention mechanisms for network equipment fault prediction. 
\cite{Bradford2021KnowledgeRF} explores Knowledge Representation in causal discovery, in enhancing constrained devices and IoT systems. However, their approach is only suited for small scenarios and not clearly applicable for complex dynamic systems.
\cite{liang2024partial} used Partial Information Decomposition to differentiate information types in Causal Discovery for wireless IoT, but does not account for dynamic environments that require runtime adaptation.
\cite{pham2024root} presents a comprehensive evaluation of causal inference-based Root Cause Analysis (RCA) methods for microservice systems using metrics data. The authors assess various causal discovery algorithms and RCA techniques on a combination of synthetic and real-world microservice system datasets, highlighting that no single method performs optimally across all scenarios and that current synthetic datasets may not accurately reflect real-world system behavior.
\cite{wang2024mrca} proposes a metric-level RCA approach that utilizes multi-modal data from traces, logs, and metrics to achieve a more comprehensive system understanding and localize root causes. The authors employ methods such as ranking abnormal services based on multi-modal data and building causal graphs of abnormal metrics with a reward mechanism to identify metric-level root causes efficiently and accurately.
The application of causal discovery for IoT remains largely limited to anomaly detection and intrusion detection~\cite{liang2024partial, gad2025toca, zheng2024causal, zeng2025causal, bhaskara2023causal}, while its use in self-adaptive systems (i.e., systems that use the discovered causal structure to adapt themselves) remains sparse~\cite{kashikar2025causally}.

Existing approaches leverage Causality at the network-layer as well as in small-scale static IoT systems. To the best of our knowledge, this is the first work that  leverages Causal RL through an automated pipeline to analyze the key causes of QoS violations in dynamic IoT systems and take tailored adaptation actions.

\section{Limitations}
\label{chap:crafter/sec:limitations}
While \oursystem demonstrates promising results for causality-based self-adaptation in autonomous IoT systems, several limitations must be acknowledged:
\newline
\noindent \textbf{Dependence on dataset quality.} \oursystem relies on IoT performance metrics datasets to perform causal discovery at design-time. As with all data-driven approaches, the quality, diversity, and representativeness of the collected IoT metrics dataset directly affect the accuracy of the generated causal graph. If important variables are missing, poorly measured, or insufficiently varied, the resulting causal model may be incomplete or biased. Although \oursystem can incorporate domain knowledge through expert elicitation or ontologies, fully mitigating data sparsity and hidden confounders remains challenging in complex real-world deployments.
\newline
\noindent \textbf{Scalability of causal discovery algorithms.} Although the PC-based causal discovery used in CRAFTER scales reasonably for the evaluated scenarios, causal discovery algorithms are known to face scalability issues as the number of variables and the dataset size increase. Large datasets collected from large-scale smart environments with thousands of devices, metrics, and contextual variables may require dimensionality reduction, hierarchical causal modeling, or incremental causal discovery techniques, which are not yet integrated into \oursystem.
\newline
\noindent \textbf{Stability of RL-based adaptation.} \oursystem employs Reinforcement Learning for runtime adaptation, which introduces training overhead and potential instability during learning phases. Although causal graphs contribute to reducing the state space and improve convergence, online retraining or frequent environment changes may still incur adaptation delays or transient QoS degradation. Furthermore, the current implementation does not explicitly enforce safety constraints during exploration, which may be required in critical IoT deployments.
\newline
\noindent \textbf{Simulation-based evaluations.} Our experimental evaluation relies primarily on emulated environments and realistic datasets. While this allows controlled experimentation and reproducibility, real-world deployments may exhibit additional sources of noise, failures, and non-stationarity that are difficult to fully emulate. As a result, a sim-to-real gap may exist when deploying \oursystem in production IoT systems, particularly for safety-critical or latency-sensitive applications.

\section{Conclusion and Future Work}
\label{chap:crafter/sec:conclusion}
This paper presents \oursystem to enable causal-based self-adaptation for autonomous IoT systems.
We propose a formal approach based on Causal Discovery for identifying the causes of performance degradation in IoT systems.
This results in causal models that are used for modelling states for an Adaptation Agent that is trained to take adaptation decisions in dynamic environments.
The experimental results show that \oursystem achieves a performance improvement of $25\%$ over state-of-the-art RL approaches used for self-adaptation.

Our future work includes adding a Model Agnostic Meta-Learning (MAML) component for predicting future states of the system, in terms of context and QoS. 
We shall also investigate causal inference techniques for quantifying causal effects among the different variables.
Finally, we will evaluate \oursystem in real-life deployments.

\begin{acks}
This work has received funding from the European Union’s Horizon Europe research and innovation actions under grant agreements No. 101135775 (PANDORA) and No. 101168560 (CoEvolution).
\end{acks}

\bibliographystyle{ACM-Reference-Format}
\bibliography{bibliography}

\end{document}